# ICADx: Interpretable computer aided diagnosis of breast masses

Seong Tae Kim[a*], Hakmin Lee[a*], Hak Gu Kim[a], and Yong Man Ro[a]

[a]Image and Video Systems Lab., School of Electrical Engineering, KAIST, South Korea

**ABSTRACT**

In this study, a novel computer aided diagnosis (CADx) framework is devised to investigate interpretability for classifying breast masses. Recently, a deep learning technology has been successfully applied to medical image analysis including CADx. Existing deep learning based CADx approaches, however, have a limitation in explaining the diagnostic decision. In real clinical practice, clinical decisions could be made with reasonable explanation. So current deep learning approaches in CADx are limited in real world deployment. In this paper, we investigate interpretability in CADx with the proposed interpretable CADx (ICADx) framework. The proposed framework is devised with a generative adversarial network, which consists of interpretable diagnosis network and synthetic lesion generative network to learn the relationship between malignancy and a standardized description (BI-RADS). The lesion generative network and the interpretable diagnosis network compete in an adversarial learning so that the two networks are improved. The effectiveness of the proposed method was validated on public mammogram database. Experimental results showed that the proposed ICADx framework could provide the interpretability of mass as well as mass classification. It was mainly attributed to the fact that the proposed method was effectively trained to find the relationship between malignancy and interpretations via the adversarial learning. These results imply that the proposed ICADx framework could be a promising approach to develop the CADx system.

## 1. INTRODUCTION

In real clinical practice, radiologists usually read medical images and find the abnormalities to determine that the patient has cancer. Normally, clinical decisions are made with reasonable descriptions which are recorded in clinical reports. For the case of breast cancer diagnosis, the Breast Imaging Reporting and Data System (BI-RADS) is defined by the American College of Radiology (ACR) and widely used as a standardized method to record and communicate the abnormalities.[1]

Recently, a deep learning technology has dramatically succeed in various applications such as image/video recognition,[2–4] biometrics,[5–8] image generation,[9–11] and medical image analysis[12–15] as well. Deep learning approaches have achieved impressive accuracies in computer-aided detection (CADe) and computer-aided diagnosis (CADx) on various modalities.[12,13,15] The use of the current deep learning approaches for CADx in real world is limited due to the lack of interpretability. So, the interpretability of the decisions made by the deep learning approach needs to be investigated for real world deployment.

In this study, a novel interpretable CADx (ICADx) framework is devised to overcome the aforementioned limitation. The proposed framework is designed to make a diagnosis decision and to provide interpretation of the decision with a standardized description (*e.g.*, BI-RADS). Note that radiologist records patients abnormalities based on BI-RADS in breast cancer diagnosis. To effectively learn the relationship between malignancy and BI-RADs description, a generative adversarial network framework is devised. The generative adversarial network framework consists of a synthetic lesion generative network and an interpretable diagnosis network. To the best of our knowledge, the approach for interpreting the decision of deep learning with a standardized medical description has not been reported in the area of CADx. Moreover, the proposed framework could provide visual interpretation with the synthetic lesion generative network. In the experiments, the effectiveness of the proposed method has been validated on the public mammogram database. Experimental results showed that the proposed method could effectively provide interpretation of diagnosis as well as mass classification. It means that the proposed deep network effectively learns the relationship between malignancy and standardized medical descriptions through adversarial learning. The visual interpretation on diagnosis could provide the reliability of the diagnostic decision and interpretation.

---

*Both authors equally contributed to this manuscript.

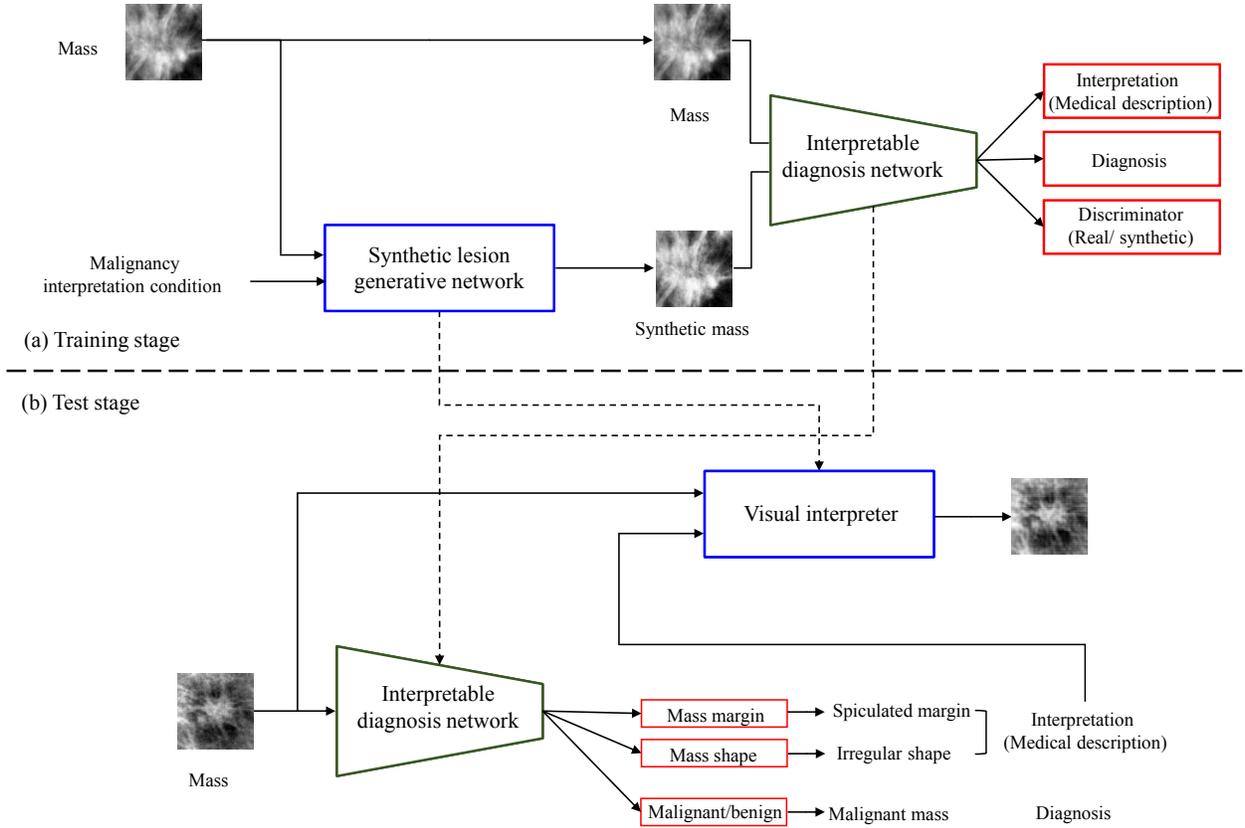

Figure 1. Framework of the proposed interpretable CADx.

## 2. PROPOSED INTERPRETABLE CADX

An overall structure of the proposed interpretable CADx framework is shown in Figure 1. As shown in the figure, the framework largely consists of 1) a synthetic lesion generative network and 2) an interpretable diagnosis network. In the training stage, the generative network and the diagnosis network compete in a two-player minimax game.[9] The generative network tries to synthesize realistic-looking lesions from malignancy interpreting conditions (*e.g.*, margin and shape of masses). The diagnosis network tries to classify malignant masses and benign masses and discriminate real and synthetic masses. Moreover, the diagnosis network tries to predict medical description. With an alternative training in minimax two-player game, the generative network and the diagnosis network can be boosted. In the test stage, the interpretable diagnosis network diagnoses mass with interpretation (medical description). The synthetic lesion generative network is used for providing visual interpretation of diagnosis. Details and learning procedure are described in the following subsections.

### 2.1 Synthetic Lesion Generative Network

In order to learn the relationship between malignancy and BI-RADS description, the synthetic lesion generative network is devised to boost the interpretable diagnosis network through adversarial learning. To synthesize the mass with malignancy interpretation condition, the proposed synthetic lesion generative network takes an input image $\mathbf{x} \in \mathbb{R}^{64 \times 64}$ and malignancy interpretation conditions (*i.e.*, margin label code $\mathbf{y}^m \in \mathbb{R}^{N_m}$ and shape label code $\mathbf{y}^s \in \mathbb{R}^{N_s}$) as shown in Figure 2. $N_m$ and $N_s$ denote the number of mass margin and the number of mass shape. Each label code is passed through two fully connected layers with leaky ReLU.[16] Through the fully connected layers, each label code is embedded to a 256-dimensional vector and 4096-dimensional vector. The 4096-dimensional vector is transformed to a spatial map (*i.e.*, label channels). The input image ($\mathbf{x}$) and label channels ($\mathbf{I}_m \in \mathbb{R}^{64 \times 64}$ and $\mathbf{I}_s \in \mathbb{R}^{64 \times 64}$) are concatenated and induced to the synthetic lesion generative network.

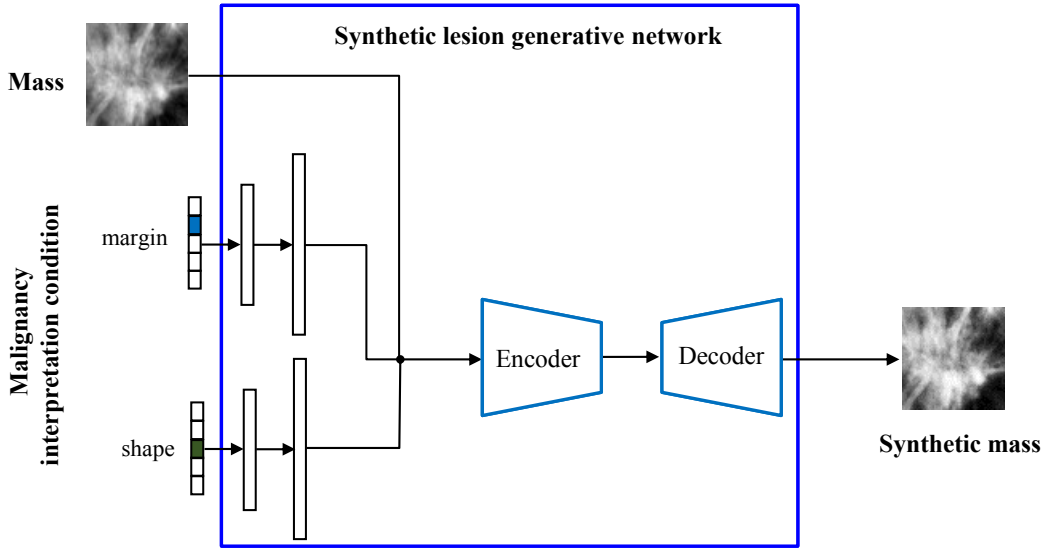

Figure 2. Detail structure of the synthetic lesion generative network.

A U-Net structure,[17] which consists of encoder and decoder with skip connection, is used as to synthesize masses. The U-Net structure connects low layers in encoder and high layers in decoder by skip connection. The U-Net structure allows the generative network to maximize the utilization of low-level information to elaborately generate the masses. The noise is provided in the form of the dropout[18] on the first two layers in the decoder of the generator, which enables the generative network to make synthesized masses with variability.

## 2.2 Interpretable Diagnosis Network

The interpretable diagnosis network is designed to conduct diagnosis (*i.e.*, classification of the malignant and benign masses) and interpret the diagnostic decision with medical descriptions. As medical descriptions, margin and shape of masses are used in this study. Margin of masses and shape of masses which are representative characteristics of masses defined by BI-RADS. Figure 3 shows the example of the diagnosis results in inference step (test stage). As shown in the figure, the proposed interpretable diagnosis network provides the explanation on diagnostic decision in terms of BI-RADS description. For that purpose, the interpretable diagnosis network is designed by a multi-task convolutional neural network (CNN). The VGG structure[19] has been modified to conduct multitask predictions (classification of malignancy, shape and margin of masses).

## 2.3 Learning Generative Adversarial Network

As explained in aforementioned subsections, ICADx network consists of two networks: the synthetic lesion generative network ($G$) and the interpretable diagnosis network ($D$). The interpretable diagnosis network which is the multi-task CNN consists of multiple components. In this paper, we have four components: $D = [D^d, D^m, D^s, D^r]$. $D^d \in \mathbb{R}^2$ is for mass diagnosis (*i.e.*, malignant and benign classification). $D^m \in \mathbb{R}^{N_m}$ is for mass margin, and $D^s \in \mathbb{R}^{N_s}$ is for mass shape. $D^r \in \mathbb{R}^1$ is for real/synthetic mass classification. For learning the interpretable diagnosis network ($D$), we define the following loss functions:

$$L_{gan}^D = -\mathbb{E}_{\mathbf{x},\mathbf{y} \sim p_d(\mathbf{x},\mathbf{y})}\left[\log D^r(\mathbf{x})\right] - \mathbb{E}_{\mathbf{x},\mathbf{y} \sim p_d(\mathbf{x},\mathbf{y}),\mathbf{m} \sim p_m(\mathbf{m}),\mathbf{s} \sim p_s(\mathbf{s})}[\log(1 - D^r(G(\mathbf{x},\mathbf{m},\mathbf{s})))], \quad (1)$$

$$L_{diagnosis}^D = -\mathbb{E}_{\mathbf{x},\mathbf{y} \sim p_d(\mathbf{x},\mathbf{y})}\left[\log D_{\mathbf{y}^d}^d(\mathbf{x}) + \log D_{\mathbf{y}^m}^m(\mathbf{x}) + \log D_{\mathbf{y}^s}^s(\mathbf{x})\right], \quad (2)$$

where $\mathbf{x}$ denotes the mass image, $G(\mathbf{x},\mathbf{m},\mathbf{s})$ denotes the synthetic lesion generated by the synthetized lesion generative network in Figure 1 (a). $\mathbf{y} = \{\mathbf{y}^d, \mathbf{y}^m, \mathbf{y}^s\}$ denotes the label of mass $\mathbf{x}$ in training data where $\mathbf{y}^d$ denotes the label for mass diagnosis (malignant/benign). $p_d(\mathbf{x},\mathbf{y}), p_m(\mathbf{m})$, and $p_s(\mathbf{s})$ are distributions of training samples for diagnosis, margin, and shape. Finally, the interpretable diagnosis network ($D$) is trained by using the following objective function:

$$\min_D L^D = L_{gan}^D + \lambda_d L_{diagnosis}^D. \quad (3)$$

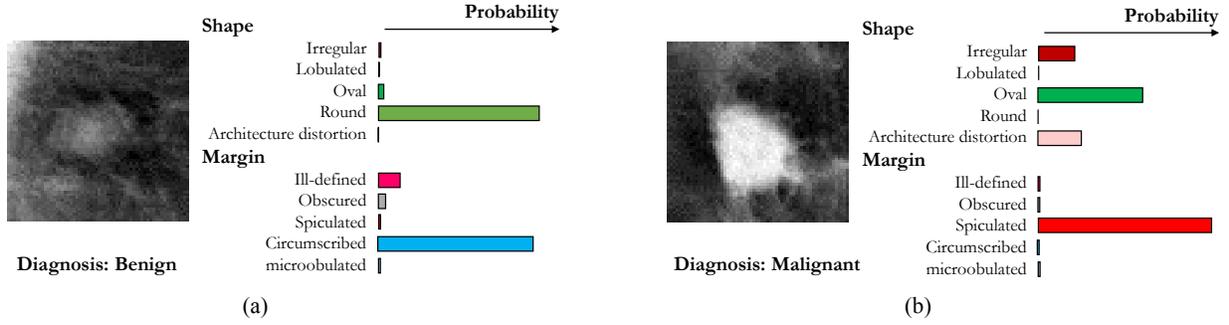

Figure 3. Example of CADx results. The proposed method could provide diagnosis results with interpretation based on radiologists terminology (margin and shape).

For learning the synthetic lesion generative network ($G$), we define the following loss functions:

$$L_{gan}^G = -\mathbb{E}_{\mathbf{x},\mathbf{y} \sim p_d(\mathbf{x},\mathbf{y}),\mathbf{m} \sim p_m(\mathbf{m}),\mathbf{s} \sim p_s(\mathbf{s})}[\log D^r(G(\mathbf{x},\mathbf{m},\mathbf{s}))], \quad (4)$$

$$L_{diagnosis}^G = -\mathbb{E}_{\mathbf{x},\mathbf{y} \sim p_d(\mathbf{x},\mathbf{y})}\left[\log D_{\mathbf{y}^d}^d(G(\mathbf{x},\mathbf{m},\mathbf{s})) + \log D_{\mathbf{y}^m}^m(G(\mathbf{x},\mathbf{m},\mathbf{s})) + \log D_{\mathbf{y}^s}^s(G(\mathbf{x},\mathbf{m},\mathbf{s}))\right], \quad (5)$$

$$L_{recon}^G = \mathbb{E}_{\mathbf{x},\mathbf{y} \sim p_d(\mathbf{x},\mathbf{y})}\left[\|\mathbf{x} - G(\mathbf{x},\mathbf{m},\mathbf{s})\|_1\right]. \quad (6)$$

The generative network is able to elaborately synthesize lesion and deceive the diagnosis network by using $L_1$ reconstruction loss as well as adversarial loss. As a result, the final objective function for training the synthetic lesion generative network ($G$) is the weighted average of each loss functions:

$$\min_G L^G = L_{gan}^G + \mu_d L_{diagnosis}^G + \mu_r L_{recon}^G. \quad (7)$$

The overall network is trained to alternatively minimize Eq. 3 and Eq. 7. With $D$ being more discriminative in classifying real vs. synthetic mass and malignancy of mass and predicting margin and shape of masses, $G$ strives to synthesize an identity-preserving lesion with the corresponding shape and margin to compete with $D$. In other words, the synthetic lesion generative network ($G$) and the interpretable diagnosis network ($D$) improve each other during alternative training.

---

**Algorithm 1:** Interpretation of decision using the synthetic lesion generative network for the visual interpreter

**Input**: $\mathbf{x}$: Mass image, $G$: Synthetic lesion generative network, $D$: Interpretable diagnosis network
**Output:** $\hat{\mathbf{x}}$: Visually interpretable synthesized image, $\mathbf{I}$: Interpretation of diagnostic decision

1. *Conduct diagnosis on the mass image using the interpretable diagnosis network*
- Estimated mass diagnosis (malignant/ benign): $\hat{\mathbf{y}}^d = D^d_{\mathbf{y}^d}(\mathbf{x})$
- Estimated medical description: mass margin $\hat{\mathbf{y}}^m = D^m_{\mathbf{y}^m}(\mathbf{x})$, mass shape $\hat{\mathbf{y}}^s = D^s_{\mathbf{y}^s}(\mathbf{x})$
2. *Generate visually interpretable synthesized image*: $\hat{\mathbf{x}} = G(\mathbf{x}, \hat{\mathbf{y}}^m, \hat{\mathbf{y}}^s)$
3. *Compare the visually interpretable synthesized image $\hat{\mathbf{x}}$ with the mass image $\mathbf{x}$*
- If $\|\hat{\mathbf{x}} - \mathbf{x}\| \leq \delta$ : $\mathbf{I}$ = diagnostic decision is interpretable ($\delta$ denotes the threshold for interpretation)
  Otherwise: $\mathbf{I}$ =diagnostic decision is not interpretable

---

## 2.4 Visual Interpreter

To provide visual interpretation on the diagnostic results, the synthetic lesion generative network is used for visual interpreter. Note that the synthetic lesion generative network tries to synthesize mass with respect to the malignancy conditions. As a result, the synthesized mass for test image from the true label codes is similar to the original test image. However, the synthesized image from the false label codes (*i.e.*, different label information for test image) makes a large difference between the synthesized image and the original test image. Based on these observations, the synthetic lesion generative network could also be utilized for providing the visual information on the medical description in test stage as shown in Figure 1. The pseudocodes for visual interpretation using the synthetic lesion generative network are given in Algorithm 1. In Section 3.4.2, we discuss the usage of the synthetic lesion generative network for visual interpreter by showing the example cases.

## 3. EXPERIMENTS AND RESULTS

### 3.1 Dataset

In our experiments, public mammogram dataset named DDSM[20] was used. In DDSM, mammogram images were digitized by different scanners with different resolutions.[20] For data consistency, mammograms scanned by Howtek 960 were selected from the DDSM dataset because a large number of cases were scanned by Howtek 960. A total of 1,088 region of interests (ROIs) which were malignant or benign masses were cropped based on the radiologists annotations for the experiments. Among them, 454 ROIs were malignant cases and 634 ROIs were benign cases. For interpreting diagnosis decision, shape of masses and margin of masses, were selected from BI-RADS descriptions because these are related with decision of malignant and benign masses. Figure 4 shows detail information of the dataset. As shown in the figure, various masses found in clinical practice were included in the dataset. Both $N_m$ and $N_s$ were 5 in this study. For evaluation purpose, 5-folds cross-validation was conducted. In each test, 20% of the dataset was set aside as test sample and remaining 80% of the dataset was used for training the deep network. There were no subject-overlap between training and test dataset. Finally, the performance was reported by averaging 5-folds results.

### 3.2 Detail Architecture and Experimental Settings

For the synthetic lesion generative network, the U-Net structure[17] was used. Both the number of convolution layer in encoder and decoder of the generator was six, respectively. The number of channels of convolution layers in encoder were 64, 128, 256, 512, 512, and 512 with filter size of 4×4. The decoder had six convolutional layers with 512, 512, 512, 256, 128, and 64 channels with filter size of 4×4.

For the interpretable diagnosis network, the VGG-like structure[14, 19] was used. The number of the convolutional layer was 13 and the final convolutional layer was flattend and passed through two fully connected layers. First fully connected layer had 1,024 units and the last fully connected layer had 13 units. The 13-dimensional vector sliced into four components as mentioned in subsection 2.3. The size of mini-batch was set to 16, and Adam optimizer was used with a learning rate of 0.0002 and an exponential decay with momentum (set to 0.99). The parameters of $\lambda_d, \mu_d$, and $\mu_r$ in Eq. 3 and Eq. 7 were set to 0.1, 10, and 100, respectively.

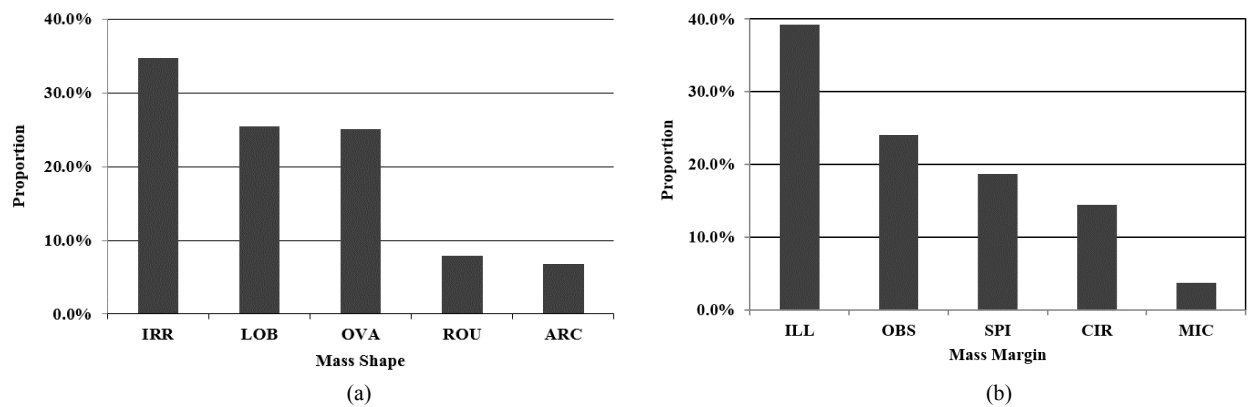

Figure 4. Information of dataset in this study. (a) Distribution of mass shapes, IRR: irregular, LOB: lobulated, OVA: oval, ROU: round, ARC: architectural distortion. (b) Distribution of mass margins, ILL: ill defined, OBS: obscured, SPI: spiculated, CIR: circumscribed, MIC: microlobulated.

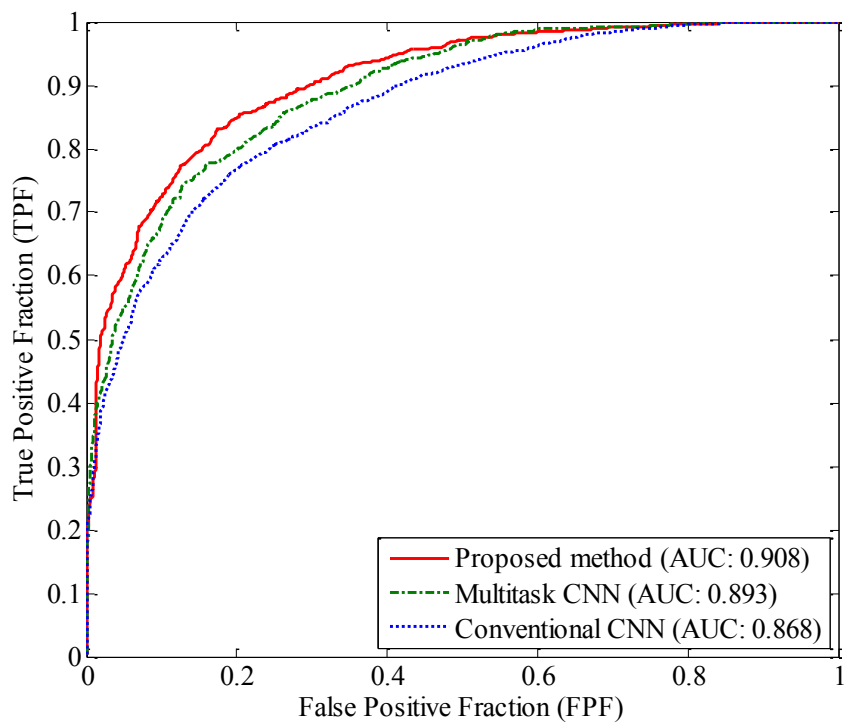

Figure 5. Comparison of ROC curves obtained from the proposed method, multitask CNN, and convetional CNN.

## 3.3 Assessment of Diagnosis

To verify the usefulness of the proposed method, comparative experiments were conducted. For the comparison, we built conventional CNN which was based on VGG[14,19] and multitask CNN. The multitask CNN was implemented on top of the conventional CNN with multitask predictions including malignancy, shape, and margin of masses. The conventional CNN and multitask CNN networks were initialized with parameters trained on large scale image classification dataset through transfer learning and re-trained on the training dataset of this study. Figure 5 shows the ROC curves[21] for evaluating diagnosis performance. As shown in the figure, the area under the ROC curve (AUC) of 0.908 was achieved by the proposed method. The AUCs of the multitask CNN and the conventional CNN were 0.893 and 0.868, respectively. The proposed method achieved the highest performance compared with the multitask CNN and the conventional CNN. It was mainly attributed to the fact that the interpretable diagnosis network was trained with the synthetic lesion generative network in the adversarial learning strategy. The competition with the generative network improved the performance of the diagnosis network.

Table 1. Comparisons of the prediction accuracy for the diagnosis results in terms of the shape and margin of masses.

| | Performance measurement | |
| --- | --- | --- |
| | **Accuracy of shape prediction** | **Accuracy of margin prediction** |
| Proposed method | 71.6% | 70.6% |
| Multitask CNN | 66.1% | 61.0% |

### 3.4 Assessment of Interpretation

#### 3.4.1 Assessment of medical description

To access the performance of diagnosis prediction with interpretation, the comparative experiment was conducted. Table 1 shows the diagnosis prediction accuracy with the interpretation in terms of medical description (BI-RADS). As shown in the table, by learning the proposed method with adversarial learning strategy using Eq. 3 and Eq. 7, the proposed method achieved higher prediction accuracy in terms of medical description (BI-RADS) compared with multitask CNN. It was mainly due to the reason that the proposed method could learn the relationship between diagnostic result and medical description. The proposed method embedded medical description (*i.e.*, malignancy interpretation conditions) to image domain and learned the relationship between medical description and malignancy of masses through the adversarial learning strategy. As a result, the interpretable diagnostic network was trained to provide accurate interpretation on diagnostic decision. In order to show the effectiveness of the proposed method, the masses synthesized from the proposed synthetic lesion generative network are represented as shown in Figure 6. As shown in the figure, the proposed method elaborately synthesized the masses with respect to the malignancy interpretation conditions.

#### 3.4.2 Visual Interpretation at test stage

As explained in subsection 2.4, the synthetic lesion generative network could be used for visual interpreter in test stage. Figure 7 shows the example of the masses and the synthesized image. Figure 7 (b) and (e) show the synthesized from the true margin and shape label codes. For figure 7 (b), the visually interpretable synthesized image was generated by using the label codes of spiculated margins and irregular shape. For figure 7 (e), the image was generated by using the label codes of circumscribed margins and round shape. Due to the reason that the synthetic lesion generative network was trained to synthesize masses as realistic as possible, if the test image was put in the generative network with true label codes, the synthesized mass was similar to the original test image. On the other hand, if the test image was put in the generative network with false label codes, the synthesized image was very different from the original test image. Figure 7 (c) was generated by obscured margins and oval shape which were frequently observed in benign masses. In other words, if the interpretable diagnosis network make wrong decision on figure 7 (a) to benign masses, the visual interpreter could not reconstruct mass image as in figure 7 (c). In the same way, figure 7 (f) was generated by ill-defined margins and irregular shape which were frequently observed in malignant masses. Based on these observations, the synthetic lesion generative network could be used as the visual interpreter for diagnostic decision and BI-RADS description. If the difference of the visually interpretable synthesized image and the test mass image is larger than pre-defined threshold $\delta$, the decision of CADx is not interpretable.

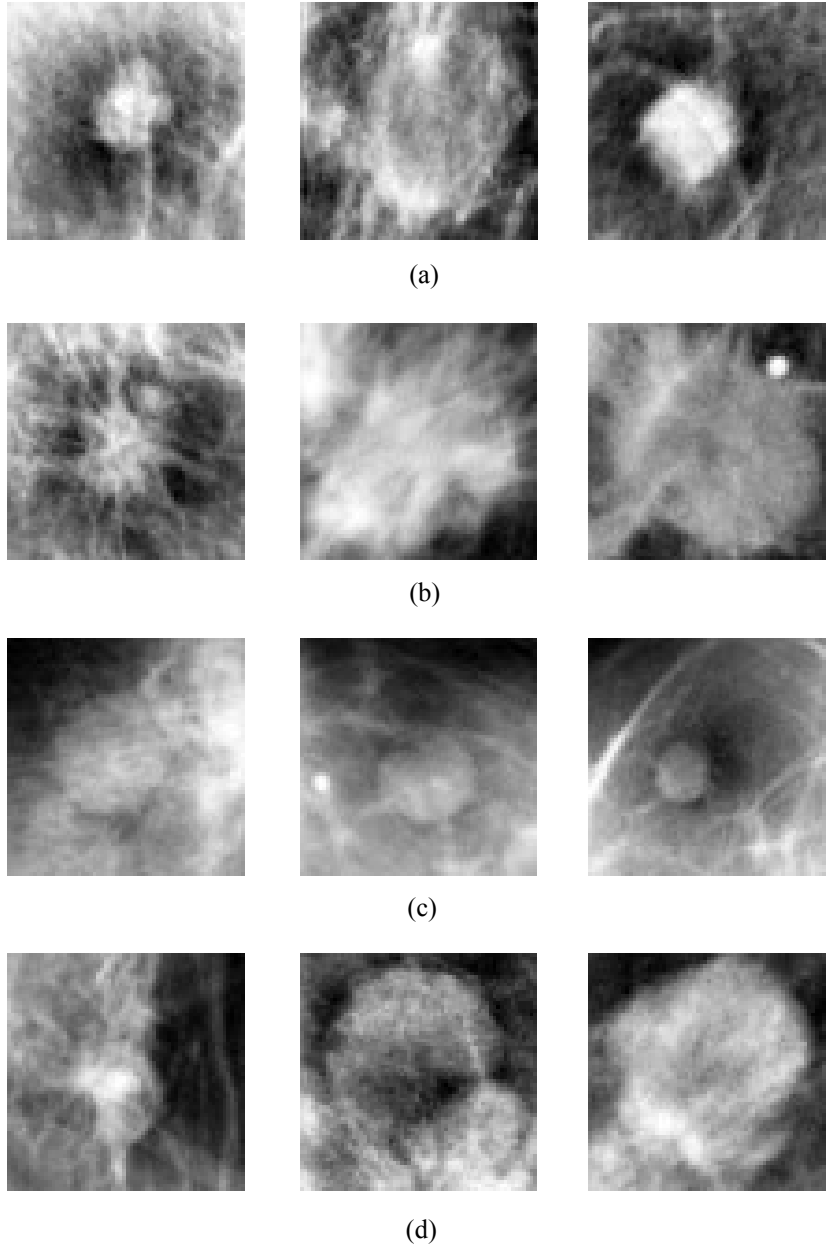

Figure 6. Example images synthesized by the proposed synthetic lesion generative network. (a) Synthesized masses with spiculated margins. (b) Synthesized masses with ill defined margins. (c) Synthesized masses with circumscribed margins. (d) Synthesized masses with obscured margins.

## 4. CONCLUSIONS

In this study, the interpretable CADx framework has been proposed to provide the diagnostic decision with interpretation in terms of medical descriptions (BI-RADS). To effectively learn the relationship between malignancy and medical description, the proposed method was designed with the generative adversarial network. The generative adversarial network consisted of the synthetic lesion generative network and the interpretable diagnosis network which were improved via the adversarial learning. Comparative experiments were conducted to validate the effectiveness of the proposed method. Experimental results showed that the proposed method effectively interpreted the diagnosis decision in terms of BI-RADS. Moreover, the synthetic legion generative network could be used as visual interpreter for diagnostic results. These results imply that the proposed ICADx could be a promising approach to develop the CADx system. As a future work, we have a plan to increase the number of interpretations such as density and subtlety. The interpretable deep learning approach can also be extended to CADx and CADe for other modalities such as digital breast tomosynthesis[22,23] and ultrasound.[24]

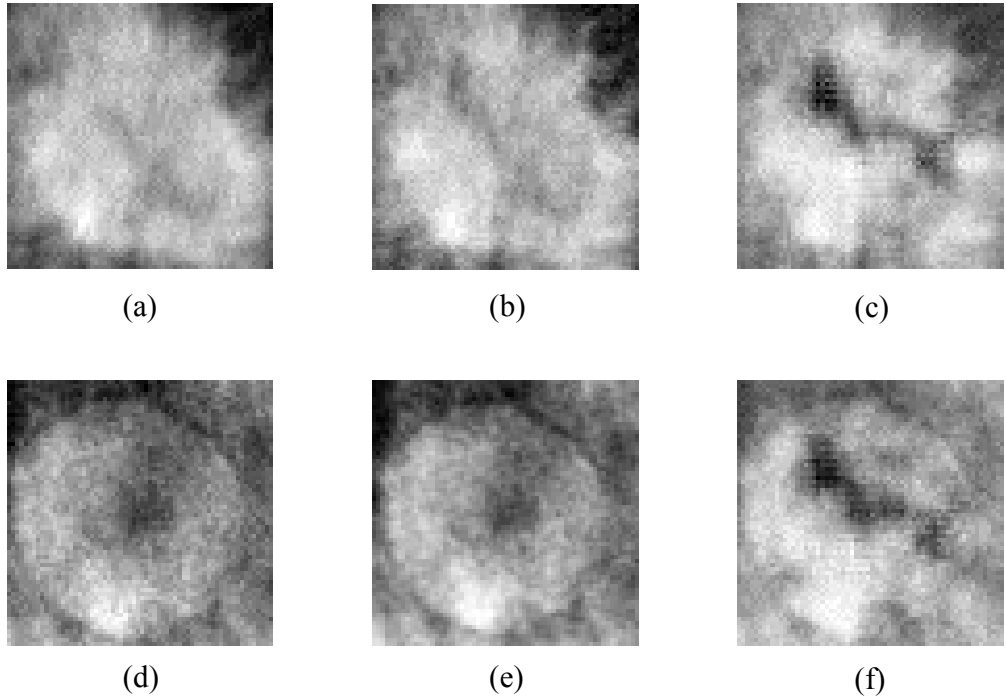

Figure 7. Visual interpretation on diagnostic results. (a) Example of malignant mass with spiculated margins and irregular shape. (b) Visually synthesized lesion with true label codes. (c) Visually synthesized lesion with false label codes. (d) Example of benign mass with circumscribed margins and round shape. (e) Visually synthesized lesion with true label codes. (f) Visually synthesized lesion with false label codes.

## ACKNOWLEDGMENTS

This work was supported by Institute for Information & communications Technology Promotion (IITP) grant funded by the Korea government (MSIT) (No. 2017-0-01778, Development of Explainable Human-level Deep Machine Learning Inference Framework).